# Ideogram Based Chinese Sentiment Word Orientation Computation

XIANG Luojie
Department of Information and Communications Engineering
University of Electro-Communications
Tokyo, Japan

*Abstract*—**This paper presents a novel algorithm to compute sentiment orientation of Chinese sentiment word. The algorithm uses ideograms which are a distinguishing feature of Chinese language. The proposed algorithm can be applied to any sentiment classification scheme. To compute a word's sentiment orientation using the proposed algorithm, only the word itself and a precomputed character ontology is required, rather than a corpus. The influence of three parameters over the algorithm performance is analyzed and verified by experiment. Experiment also shows that proposed algorithm achieves an F Measure of 85.02% outperforming existing ideogram based algorithm.**

## I. INTRODUCTION

Sentiment analysis aims to extract subjective information from mass corpus. It is widely used in fields such as business review analysis, spam filtering etc. A sentiment word is a word that expresses human emotion or opinion which is said to have sentiment orientation. Sentiment orientation expresses whether the sentiment is positive or negative or even more sophisticated sentiment classification such as happy, sad, fear or angry. Sentiment word orientation computation is the basis of higher level sentiment analysis such as that of sentence level or text level. Accuracy of sentiment word orientation computation is crucial to the higher level sentiment analysis.

Previously, several works dealt with Chinese sentiment word orientation computation. Most of the methods can be classified into two categories: statistical approach and machine learning approach [10]. For statistical approach, the sentiment orientation of words is inferred based on the similarity between the words to be measured and the reference word pairs (RWP) [10]. The similarity can be computed by co-occurrence frequency, point-wise mutual information or contextual information. Zhu et al. computes the similarity based on the HOWNET [1, 2], a knowledge base where the concepts are represented by words and the relationship between the concepts is studied. Wang et al. add synonym evaluation to Zhu's method [3]. Du et al. build undirected graph based on the similarity between words and transform the problem into function optimization [4]. Zhao et al. define the semantic distribution state of a word by its context feature then utilize a graph approach to compute word sentiment orientation by the distributional similarity [5]. The machine learning approach specifies some features and maintains a decision algorithm such as Naïve Bayes, SVM, etc. to classify the words [7, 8, 9].

A distinguishing feature of Chinese from other languages like English is the ideogram. Most Chinese characters are ideograms which have their own meaning. Very few Chinese characters are phonograms which represent the phonemes. Ideograms constituting a word contribute to its meaning and emotional orientation. For example, *bei* means sad and most words containing this character exhibit a sad emotion. Thus, the ideogram feature should be utilized in the sentiment orientation computation of Chinese sentiment words.

Ku et al. introduced an ideogram-based Chinese sentiment word orientation computation method [6]. However, they only use the frequency of occurrence when computing a character's sentiment information and deals with the sentiment classified into only positive and negative.

This paper proposes a novel algorithm based on ideograms. The proposed algorithm can be applied to any sentiment classification scheme. It does not require a corpus to compute the sentiment orientation of a word. It only requires the word itself and a precomputed character ontology, a set of characters annotated with the sentiment information. A system which can establish a character ontology and compute the Chinese sentiment word orientation is realized by implementing the proposed algorithm. The influence of three parameters over the algorithm performance is analyzed and verified by experiments using the system. The experiments also show that the proposed algorithm achieves an F Measure of 85.02% outperforming existing ideogram based algorithms.

## II. IDEOGRAM BASED ALGORITHM

The algorithm utilizes precomputed sentiment information of characters to compute the sentiment orientation of words. First, the sentiment information of some characters is calculated using a training word list, a list of sentiment words manually selected and annotated with sentiment information. These characters together with their computed sentiment information form the character ontology. Then, the sentiment orientation of new unmarked words is computed using the sentiment information of characters in the character ontology. The algorithm consists of two parts: character ideogram evaluation which computes the sentiment information of characters from the training word list and word ideogram

The author is sponsored by JASSO scholarship.

evaluation which computes the sentiment orientation of words from the sentiment information of characters in the character ontology.

### *A. Notations*

Let $c$ denote a character and $w$ denote a word. If $c \in w$, $c$ is a character in word $w$. A sentiment category is a subset of sentiment that denotes a certain emotion or opinion, such as happy, sad etc. Let $S$ denote a sentiment category. If $c \in S$, character $c$ exhibit the sentiment orientation denoted by the sentiment category $S$. Similarly, $w \in S$ means word $w$ exhibit the sentiment orientation denoted by the sentiment category $S$.

Suppose the sentiment is classified into n categories: $S_1, S_2, \ldots, S_n$. The sentiment information of a character $c$ is represented by a sentiment vector $\boldsymbol{C} = (P_1, P_2, \ldots, P_n)$ where $P_i$=Pr($c \in S_i$) and $\sum_{i=1}^{n} P_i = 1$. Similarly, the sentiment information of a word $w$ is represented by $\boldsymbol{W} = (P_1, P_2, \ldots, P_n)$ where $P_i$=Pr($w \in S_i$) and $\sum_{i=1}^{n} P_i = 1$. Sentiment orientation of a character or a word is defined as the sentiment category $S_i$ of its sentiment vector with the largest probability $P_i$.

### *B. Character Ideogram Evaluation*

This part of algorithm computes the sentiment vector of characters from a training word set $V$ and decides whether they should be added into the character ontology.

The training word set $V$ consists of a set of manually selected words that exhibit strong sentiment orientation. It should be guaranteed that, for each sentiment category $S_i$, the number of words in the training word set $V$ belonging to $S_i$ should be larger than a minimum acceptable number. The words in $V$ are manually annotated with sentiment vectors. The sentiment vector of a word in $V$ is decided by the following rule: $P_i = 1$ if $w \in S_i$ and $P_i = 0$ if $w \notin S_i$. An example is shown in Table I where sentiment is classified into happy, angry, sad and fear.

Upon the computation of the sentiment vector $\boldsymbol{C}$ of a single character $c$, firstly a word set

$$V_c = \{w | c \in w,\ w \in V\} \tag{1}$$

is retrieved. Then, the sentiment vector $\boldsymbol{C}$ is calculated by:

$$\boldsymbol{C} = \frac{1}{|V_c|} \sum_{w_i \in V_c} \boldsymbol{W}_i \qquad (V_c \neq \emptyset). \tag{2}$$

A special case is considered: a negative before $c$. A negative is a character with a negative meaning, such as *bu* which means 'no' in English. Negatives in a word tend to reverse the original sentiment orientation of its following characters. To counteract the effect of the negative in a word $w_i$, the $\boldsymbol{W}_i$ should be adjusted when added in equation (2). Though a character after a negative is turned to the opposite sentiment orientation, sentiment isn't generally classified in such a way that each sentiment category has its opposite. The solution of this problem is giving as follows. The $\boldsymbol{W}_i$ of the $w_i$ that doesn't have a negative before $c$ is first added to $\boldsymbol{C}$ in equation (2). Then, for a $w_i$ that has a negative before $c$, the sentiment orientation of $c$ is assumed to be turned into the same sentiment orientation that $\boldsymbol{C}$ currently has. This means, the $\boldsymbol{W}_i$ of a $w_i$ that has a negative before $c$ is adjusted when added that its sentiment category $S_m$ same with $\boldsymbol{C}$'s current sentiment orientation is assigned probability 1 and the others assigned probability 0. This adjustment on $\boldsymbol{W}_i$ is only temporary.

TABLE I. EXAMPLE OF TRAINING WORD LIST ANNOTATION

| Chinese | Meaning | Sentiment Orientation | Sentiment Vector |
|---|---|---|---|
| Yu Kuai (愉快) | Enjoyment | Happy | (1,0,0,0) |
| Gu Du (孤独) | Loneliness | Sad | (0,0,1,0) |
| Kong Bu (恐怖) | Horror | Fear | (0,0,0,1) |

a. Each component of sentiment vector corresponds to sentiment category of happy, angry, sad, fear

After the sentiment vector of a character is computed, it should be decided whether that character and its sentiment vector should be added into the character ontology. Because different characters has different contribution to the sentiment orientation computation of words, some characters and their sentiment vectors are not added into the character ontology while some are added or even with adjustments on their sentiment vectors to emphasize their significance. Three things are considered during this decision.

*Key characters* exhibit strong sentiment orientation. Words containing those characters have a very high possibility to exhibit the same sentiment orientation with them. For example, *le*, which means happy, exhibits a very strong sentiment orientation in happiness. Words containing *le* (with no negatives before it) have a very high probability to exhibit a happy emotion. Due to the significance of key characters over sentiment orientation computation of words, a parameter *Upper Variance Bound* (*UVB*) is set to identify key characters. If the variance of all components of a character's sentiment vector is greater than *UVB*, it is set as a key character. Once a character is rated as a key character, its sentiment vector is adjusted in such a way that the maximum $P_i$ is augmented to 1 with the others reduced to 0 to magnify the influence of key characters over sentiment orientation computation of words. Notice that, this adjustment on the sentiment vector is permanent.

*Nonsentiment characters* exhibit no or little sentiment orientation. They are useless for word sentiment orientation computation and are not added into the character ontology. A parameter *Lower Variance Bound* (*LVB*) is set to identify the nonsentiment characters. If the variance of all components of a character's sentiment vector is smaller than *LVB*, it is rated as nonsentiment character.

*Threshold*. A low occurrence frequency of a character is likely to result in miscomputation of its sentiment vector. A parameter *Threshold* (*T*) is set to prevent this situation. Only when a character's occurrence frequency reaches $T$ can it be added into the character ontology.

### *C. Word Ideogram Evaluation*

This part of the algorithm computes the sentiment orientation of words based on the character ontology. After the training process, a set of characters $K$ annotated with sentiment vectors is stored in the character ontology. To compute the sentiment orientation of a word $w$, firstly a word set

$$K_w = \{c | c \in w,\ c \in K\} \tag{3}$$

is retrieved. Then, the sentiment vector $\boldsymbol{W}$ is calculated by:

$$\boldsymbol{W} = \frac{1}{|K_w|} \sum_{c_i \in K_w} \boldsymbol{C}_i \qquad (K_w \neq \emptyset). \tag{4}$$

Two special cases are considered.

*Key characters*. Due to the key character's dominant influence over the word sentiment orientation, the words containing key characters are directly assigned the same sentiment vector with the key character. An exception is if negatives exist before the key character which will be discussed later. For two or more key characters existing in a word at the same time, if they have the same sentiment vector directly or after some adjustment on their sentiment vectors due to the existence of the negatives, their common sentiment vector is assigned to the word otherwise the sentiment vector of the word can't be computed.

*Negatives*. If a negative exists in $w$, a similar technique is used as the solution for the negatives in the character ideogram evaluation part. In equation (4), the order of adding $\boldsymbol{C}_i$ is changed and the sentiment vectors of characters after the negative is temporarily adjusted when added. Sentiment vectors of characters before the negative will first be added into $\boldsymbol{W}$ without adjustment. Then, the sentiment vectors $\boldsymbol{C}_i$ of the characters $c_i$ behind the negative is added to $\boldsymbol{W}$ temporarily adjusted in a way that their sentiment category $S_m$ same with $\boldsymbol{W}$'s current sentiment orientation is assigned probability 1 and the other sentiment categories assigned probability 0. If there is a key character after a negative, the sentiment vectors of the characters before the negative are still added to $\boldsymbol{W}$ first. And for all characters after the negative, only key character is added into $\boldsymbol{W}$ with its sentiment vector adjusted in the same way mentioned above. If the first character of $w$ is a negative, $\boldsymbol{W}$ can't be computed.

## III. EXPERIMENT

The algorithm performance is assessed by F Measure which is the harmonic mean of the precision and recall. Precision is the number of correctly computed words divided by the number of words computed. Recall is the number of correctly computed words divided by the number of all words that should have been correctly computed.

Three parameters mentioned above affect the algorithm performance: *Threshold* ($T$), *Lower Variance Bound* ($LVB$) and *Upper Variance Bound* ($UVB$). Their influence over the algorithm performance is analyzed and then verified by the experiment below.

*Threshold (T)*. Small $T$ value results in many characters in the character ontology having a miscomputed sentiment vector due to their low occurrence frequency. This leads to the miscomputation of many words' sentiment orientation. As $T$ grows, this situation gets better resulting in an increase in the precision and becomes ignorable when $T$ reaches a reasonable large value. Meanwhile, however, another factor, the miscomputation of the word sentiment orientation due to the words containing characters with their minor meaning, becomes considerable. With an increasing $T$, the number of characters in the character ontology decreases rapidly, resulting in the number of words computed also decreasing rapidly (essentially a rapid decreasing recall). Whereas, the number of words miscomputed due to containing characters with minor meaning remains the same. Therefore, precision decreases after $T$ increases beyond a reasonable large value. But the F Measure should decrease as $T$ grows because recall decreases so rapidly covering up the precision's influence over the F Measure.

*Lower Variance Bound (LVB)*. Precision increases as $LVB$ increases for the number of nonsentiment characters decreases as $LVB$ increases causing a decrease in the miscomputed words. However, increasing $LVB$ also results in a rapid decrease in the number of words computed thus a rapid decrease in recall which covers up the precision's influence over the F Measure finally leading to a decreasing F Measure.

*Upper Variance Bound (UVB)*. When $UVB$ is small, many characters are wrongly rated as key characters, resulting in words miscomputed and the words having key characters with different sentiment vectors not computed. When $UVB$ grows, fewer characters are wrongly rated as key characters so that fewer words are miscomputed and fewer words have key characters with different sentiment vectors. However, those words having key characters with different sentiment vectors when $UVB$ is small have a high probability to be miscomputed when $UVB$ is reasonably large and they have only one or no key characters. Thus, precision remains stable. The F measure grows when $UVB$ grows due to the stable precision and the decrease of words having key characters with different sentiment vectors resulting in an increasing recall.

The summarization of the analysis above is given as follows:

1. As the *Threshold* ($T$) increases, the precision first increases and then decreases while the recall and the F measure decreases all the time.
2. As the *Lower Variance Bound* ($LVB$) increases, the precision increases while the recall and the F measure always decreases.
3. As the *Upper Variance Bound* ($UVB$) increases, the precision remains stable while the recall and F measure increases.

Thus, to obtain the best F measure, $T$ and $LVB$ should be as small as possible and $UVB$ should be as large as possible.

In the experiments, sentiment is divided into four categories: happy, angry, sad and fear. 1710 words exhibiting strong sentiment orientation in these four categories are manually selected from dictionary as the training word list. They are manually classified into the four sentiment categories and each word is annotated with a sentiment vector using the technique mentioned in Section 2. 30 word lists each having 500 words randomly selected from a sentiment word list provided by HOWNET [2] are generated as test word list. A system that can establish the character ontology and compute the word sentiment orientation based on the character ontology is developed by implementing the proposed algorithm. Test word lists are inputted into the system which will compute and output the sentiment orientation of all the words inputted. Meanwhile, the test word lists will be manually annotated with sentiment orientation which is rated as correct. The computed result and the manually annotated result will be compared to generate precision, recall and thus F Measure. Firstly, influence of the three parameters over the algorithm performance (F Measure) is experimented and then best F Measure the proposed algorithm can achieve. Experiment results verifying the analysis above are shown in Fig. 1, Fig. 2 and Fig. 3 (due to convenience all variance is multiplied by 4).

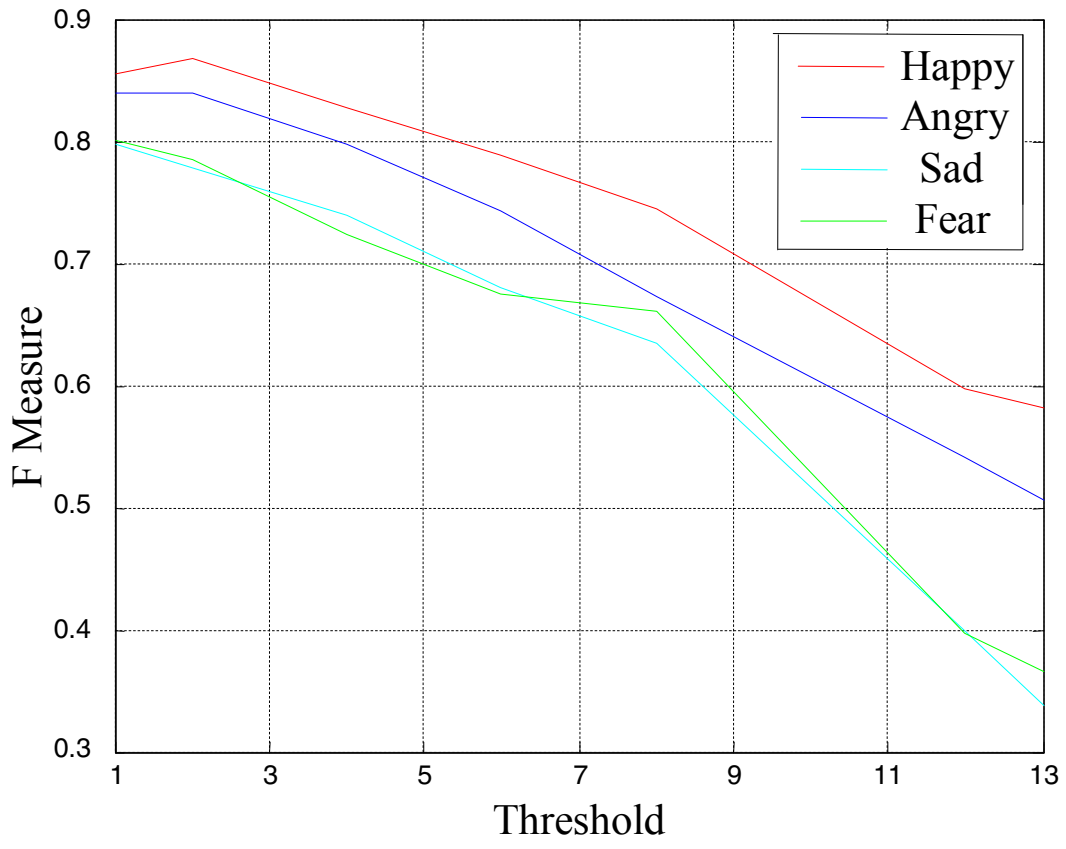

Figure 1. F Measure over *Threshold* (*LVB*=0.3 *UVB*=0.65)

According to the influence of parameters over the algorithm performance (assessed by the F Measure), the maximum F Measure (the average of the result of four sentiment categories) can be achieved when *T*=1 *LVB*=0.1 *UVB*=0.65 shown and compared to the previous ideogram based algorithm and other algorithms in Table II.

## IV. Conclusion and Future Work

Experiment shows that the proposed algorithm can achieve an F Measure of 85.02% outperforming existing ideogram based algorithm. This proves the effectiveness of the ideogram based method over the Chinese sentiment word orientation computation. The improvement on algorithm performance achieved by the proposed algorithm contributes to the higher level sentiment analysis.

Though the proposed algorithm achieved a very good result, it focuses solely on ideograms and ignores context information of words and mutual information between words. The future work is to combine this algorithm and other algorithms to fully utilize the sentiment information of a sentiment word and further improve the algorithm performance.

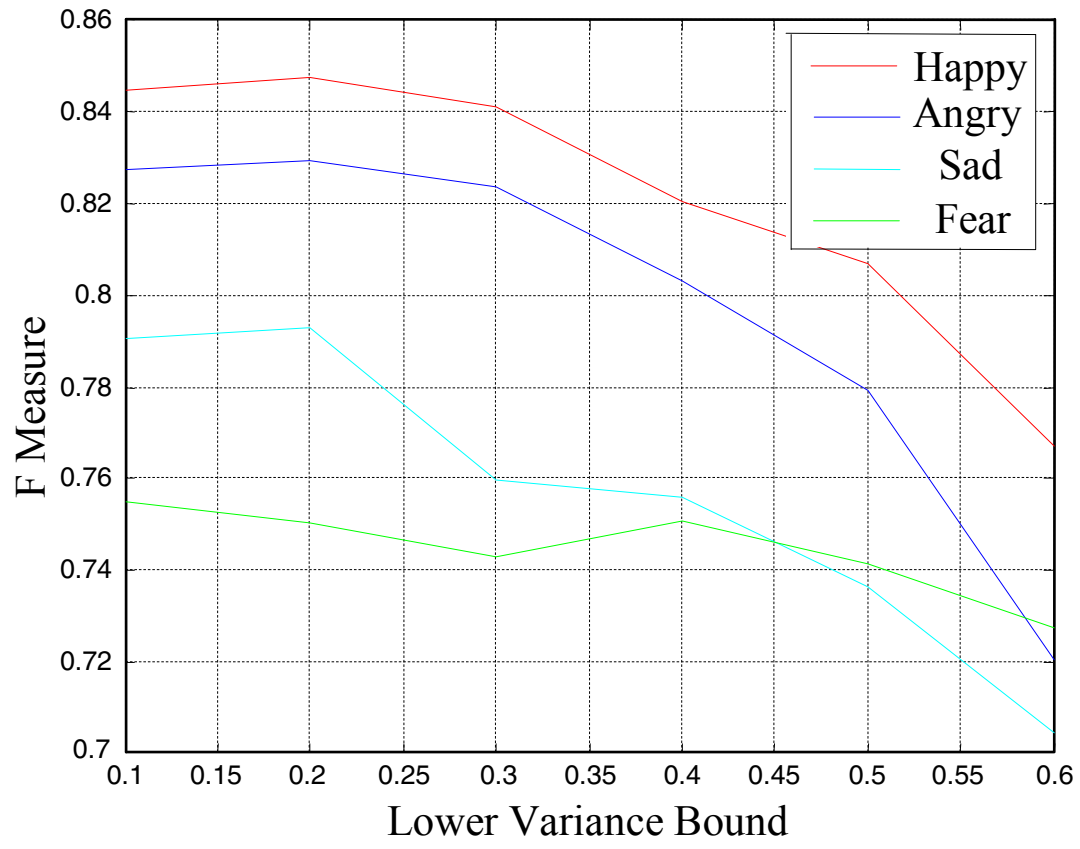

Figure 2. F Measure over *Lower Variance Bound* (*T*=6 *UVB*=0.65)

## Acknowledgment

The author thanks Professor Brian Kurkoski and Dr. Masahisa Mabo Suzuki for useful remarks and guidance.

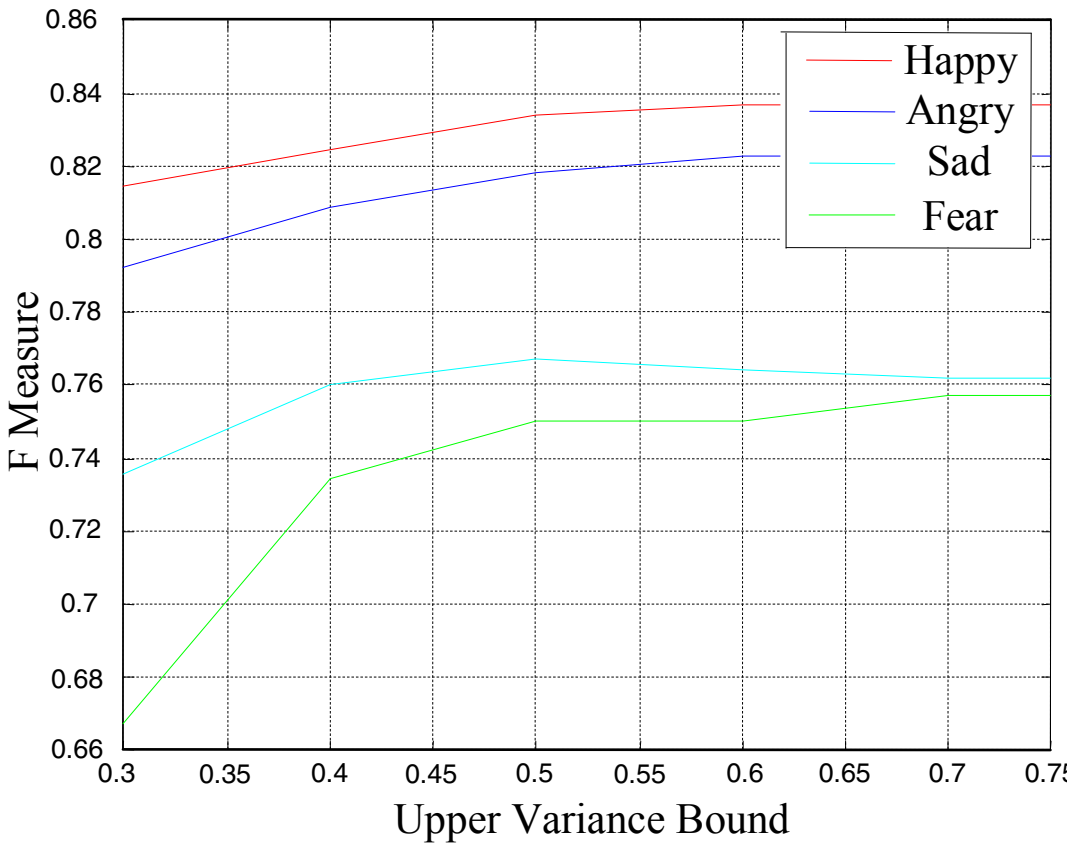

Figure 3. F Measure over *Upper Variance Bound* (*T*=6 *LVB*=0.3)

TABLE II. Algorithm Performance Comparison

| Paper | [1] | [5] | [6] | This Paper |
|---|---|---|---|---|
| Precision | 0.693 | 0.801 | 0.6106 | 0.8777 |
| Recall | 0.187 | 0.233 | 0.7942 | 0.8247 |
| F Measure | 0.295 | 0.361 | 0.6847 | 0.8502 |

a. Paper [6] is the previous ideogram based algorithm while paper [1], [5] is the work of other algorithms.

## References

[1] Zhu Yanlan, Min Jin, Zhou Yaqian, Huang Xuanjing, and Wu Lide, “Semantic orientation computing based on HowNet,” (in Chinese), *J. Chinese Inform. Process.*, vol. 20, no. 1, pp. 14-20, 2006.

[2] Dong Zhengdong and Dong Qiang. (2008, Sep. 15). *HowNet knowlegde database* (*version 2008*) [Online]. Available: http://www.keenage.com.

[3] Wang Suge, Li Deyu, Wei Yingjie, and Song Xiaolei, “A synonyms based word sentiment orientation discriminating,” (in Chinese), *J. Chinese Inform. Process.*, vol. 23, no. 5, pp. 68-74, Sep. 2009.

[4] Du Weifu, Tan Songbo, Yun Xiaochun and Cheng Xueqi, “A new method to compute semantic orientation,” (in Chinese), *J. Comput. Research and Develop.*, vol. 46, no. 10, pp. 1713-1720, 2009.

[5] Zhao Yu, Cai Wandong, Fan Na, and Li Huixian, “Computing Chinese semantic orientation via distributional similarity,” (in Chinese), *J. Xi’An JiaoTong University*, vol. 43, no. 6, pp. 33-37, Jun. 2009.

[6] Ku Lunwei, Wu Tungho, Lee Liying, and Cheng Hsinhsi, “Construction of an evaluation corpus for opinion extraction,” in *Proc. NTCIR-5 Workshop Meeting*, Tokyo, 2005, pp. 513–520.

[7] Changli Zhang, Wanli Zuo, Tao Peng, and Fengling He, “Sentiment classification for Chinese reviews using machine learning methods based on string kernel,” in *3rd 2008 Int. Conf. Convergence and Hybrid Inform. Technology*, Busan, 2008, pp. 909-914.

[8] Jianxin Yao , Gengfeng Wu , Jian Liu and Yu Zheng, “Using bilingual lexicon to judge sentiment orientation of Chinese words,” in *Proc. 6th IEEE Int. Conf. Comput. and Inform. Technology*, Seoul, 2006.

[9] Pang, B., Lee, L., and Vaithyanathan, S., “Thumbs up? Sentiment classification using machine learning techniques,” in *Conf. Empirical Methods in Natural Language Process.*, Philadelphia, 2002, pp. 79-86.

[10] Ting-Chun Peng and Chia-Chun Shih, “An unsupervised snippet-based sentiment classification method for Chinese unknown phrases without using reference word pairs,” in *2010 IEEE/WIC/ACM Int. Conf. Web Intell. and Intelligent Agent Technology*, Toronto, 2010, pp. 243-248.